\documentclass[wcp]{jmlr}

\usepackage{longtable}% for long tables
\usepackage{multirow}
\usepackage{algorithm}
\usepackage{algorithmic}
\usepackage[capitalize,noabbrev]{cleveref}
\usepackage{siunitx} % for nicely formatted numbers and units
\usepackage{caption}

\usepackage{booktabs}
\usepackage{lineno}
\makeatletter
\newcommand\blfootnote[1]{%
  \begingroup
  \renewcommand{\@makefntext}[1]{\noindent\makebox[1.8em][r]#1}
  \renewcommand\thefootnote{}\footnote{#1}%
  \addtocounter{footnote}{-1}%
  \endgroup
}
\makeatother

\makeatletter
\let\Ginclude@graphics\@org@Ginclude@graphics 
\makeatother

\title[How GAN Generators can Invert Networks in Real-Time]{How GAN Generators can Invert Networks in Real-Time}

  \author{\Name{Rudolf Herdt} \Email{rherdt@uni-bremen.de}\\
   \Name{Maximilian Schmidt} \Email{schmidt4@uni-bremen.de}\\
   \Name{Daniel Otero Baguer} \Email{otero@uni-bremen.de}\\
   \Name{Jean Le'Clerc Arrastia} \Email{leclerc@uni-bremen.de}\\
   \Name{Peter Maaß} \Email{pmaass@uni-bremen.de}\\
   \addr Center of Industrial Mathematics, University of Bremen, Germany}

\begin{document}

\maketitle

\begin{abstract}
%Critical applications, such as in the medical field, require the rapid provision of additional information to interpret decisions made by deep learning methods.
%
In this work, we propose a fast and accurate method to reconstruct activations of classification and semantic segmentation networks by stitching them with a GAN generator utilizing a 1x1 convolution. We test our approach on images of animals from the AFHQ wild dataset, ImageNet1K, and real-world digital pathology scans of stained tissue samples. Our results show comparable performance to established gradient descent methods but with a processing time that is two orders of magnitude faster, making this approach promising for practical applications.
%Our method provides comparable results to established gradient descent methods on these datasets while running about two orders of magnitude faster.
\end{abstract}

\section{Introduction}
\label{sec:intro}
%
%\blfootnote{The code is available at: https://github.com/herdtr/gan-stitching}
%
%Interpretability of neural networks and understanding what features they extract from the input are crucial, especially in critical applications like healthcare. %Additionally, the availability of this information in near real-time is often required.
%
In this work, we efficiently reconstruct the extracted features of a ResNet50 \citep{resnets} classification network trained on ImageNet1K \citep{deng_imagenet_2009} and a ResNet34 backbone of a semantic segmentation network trained on digital pathology scans of histochemically stained tissue samples, by inverting them using a pretrained, frozen generative adversarial network (GAN) \citep{NIPS2014_5ca3e9b1}. %A StyleGAN2  architecture \citep{karras2020stylegan2} is used in this role.
%
% TODO: find BigGAN iclr citation, replace arxiv one
For the ResNet50 trained on ImageNet1K, we invert it using a StyleGAN2 \citep{karras2020stylegan2} and a BigGAN \citep{brock2019large}. For the ResNet34 trained on digital pathology data, we use a custom GAN  trained on digital pathology data. The architecture is shown in \cref{apd:GAN Architecture}.
In the following, we use the term "feature extractor" for both the classification and semantic segmentation network.

%
%We observe that for most of the layers, the features learned by the feature extractor are compatible with the features learned by a GAN generator, even though the feature extractor and the GAN were trained completely unaware of each other.
%
%Only the deeper layers of a class conditional BigGAN are not compatible with a ResNet50.
%
%Further, we observe that the earlier layers of StyleGAN2 are more general, and the deeper layers become more specialized.

In order to bring the feature extractor and GAN generator together, we learn a linear transformation in the form of a 1x1 convolution to stitch them.
The 1x1 convolution is trained to map from a hidden layer of the feature extractor into a hidden layer of the GAN generator.
Utilizing this mapping, we can quickly reconstruct activations of the feature extractor, by transferring them through the convolutional connection into the GAN generator, i.e., we use the GAN generator as a decoder to invert the feature extractor.
Here, only the 1x1 convolution is being trained. Both the feature extractor and GAN generator are kept frozen.
%
%We then compare the reconstructions from the GAN with the original images using L1 loss and cosine similarity between their activations, averaged over a dataset.
%

Through the stitching, we investigate the similarity between representations learned by the GAN generators and the feature extractors, and also which combination of layers is more compatible.

\section{Related Work}
\label{sec:related}

We use model stitching \citep{7298701} to stitch a feature extractor with a GAN, which allows us to reconstruct activations of the feature extractor in real-time (on average, it takes 0.034s per image using StyleGAN2).
A 1x1 convolution is used as the stitching layer, which is methodically closely related to the work of \cite{NEURIPS2021_01ded425}. They applied a sequence of batch norm, 1x1 convolution, and batch norm to stitch two image classification networks together in a hidden layer.
In contrast, we do the stitching in order to see how similar the learned representations of hidden layers of a feature extractor are to the learned representations of hidden layers of a GAN generator by comparing the reconstructed images with the original images.
In addition, we reconstruct activations of the feature extractor by sending them through the GAN generator.

Those activations can also be reconstructed through gradient descent \citep{Mahendran2014UnderstandingDI}.
The disadvantage is that it is slow since several forward-backward passes through the network are needed.
We use gradient descent as in \cite{olah2017feature} and gradient descent without any regularization as baselines to compare our GAN method against.
Another method is to reconstruct the activations by training a decoder network to invert a feature extractor from a hidden layer \citep{AlexnetInvertDecoder}.
But such an approach makes it necessary to train a new decoder for every new layer in the feature extractor where activations should be reconstructed from.
In contrast, with our method, it is only necessary to train a new 1x1 convolution and keep the same GAN generator.
Finally, in the work of \cite{AlexnetInvertDecoder}, the decoder was trained to minimize a reconstruction error in the input space of the classification network, whereas we train the stitching layer to minimize a reconstruction error at the hidden layer where we transfer from.

\section{Methodology}
\label{sec:main}

In this section, we show an overview of our GAN-based reconstruction method and the gradient descent methods we use as a baseline. In addition, we describe the evaluation metrics used in the experiments.

\subsection{Overview}
\label{sec:overview}

\begin{figure*}[htb]
%\vskip 0.2in
\begin{center}
\centerline{\includegraphics[width=\textwidth]{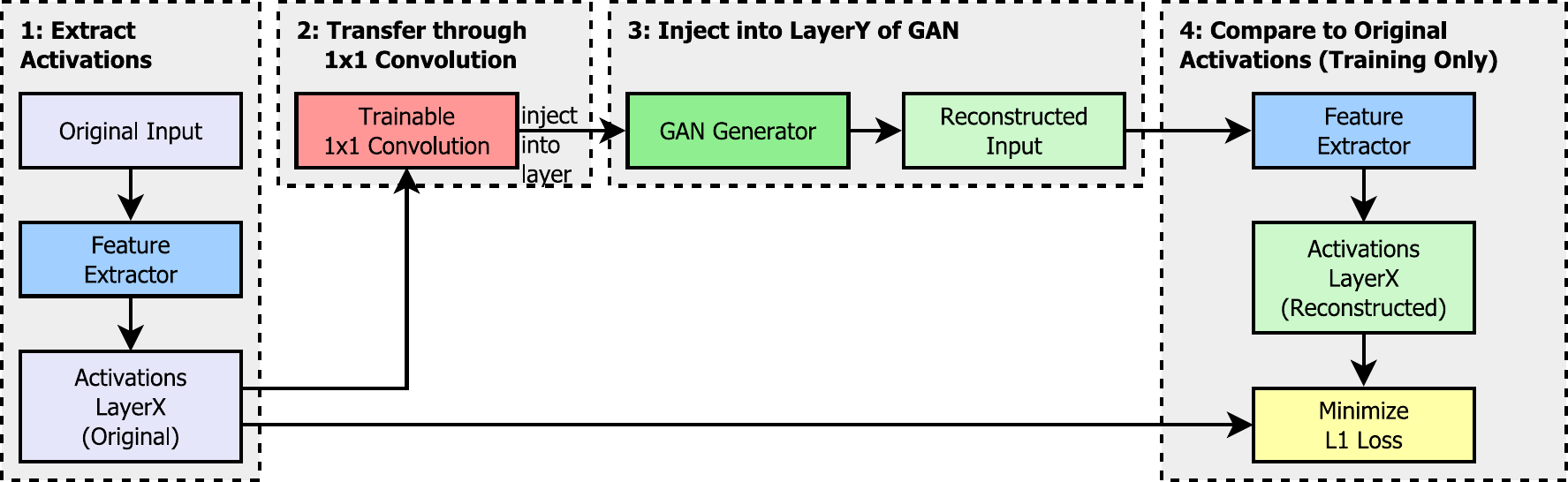}}
\caption{Usage and training process for stitching a GAN generator and a feature extractor in a hidden layer.}
\label{fig:overview}
\end{center}
\vskip -0.2in
\end{figure*}

\cref{fig:overview} shows an overview of our approach.
For combining a feature extractor with a GAN generator in a hidden layer, i.e., stitching LayerX of the feature extractor and LayerY of the GAN generator, we train a 1x1 convolution to transfer between the two layers.
After the 1x1 convolution is trained, we can invert the feature extractor at LayerX by propagating the activations from LayerY till the output of the GAN generator, as shown in steps 1 - 3 in \cref{fig:overview}.

First, the activations at LayerX are extracted. Then we transfer them through the 1x1 convolution to get them from the latent space of LayerX into the latent space of LayerY.
After that, the activations are scaled to match the spatial size of LayerY. For example, layer b32.conv0 of StyleGAN2 has a spatial size of 32x32. Therefore, before injecting our activations into that layer, we scale them to a spatial size of 32x32 via nearest neighbor scaling.
Finally, we inject the activations into LayerY of the GAN generator while generating an output with the generator from a random seed.
This means that we still get variations from the used seed or noise after the injection layer.
The output of the GAN generator is referred to as reconstructed input in \cref{fig:overview}.

For training, we need to compare the reconstructed and original input and train the 1x1 convolution to minimize a loss between the two.
To compare them, we again use the feature extractor, as shown in step 4 in \cref{fig:overview}, and optimize the 1x1 convolution to minimize the L1 loss in LayerX.
%
%We do not minimize a loss in the input layer since we expect that, as the input propagates deeper through the network, it should discard unnecessary low-level information.
%
We do not minimize the loss in the input layer, because we expect the network to increasingly discard unnecessary low-level information the deeper the input propagates through the network.
For example, individual positions of the dots on leopard fur, which the network would then be unable to reconstruct properly. Since we aim to reconstruct the activations of LayerX, we also use this layer to compute the loss.

\subsection{Evaluation}

To evaluate how good our GAN-based inversion method is, we can first look at some reconstructions, i.e., if the reconstructions do not resemble the original images, then the method has failed for this combination of layers.
To get a more comparable quantification of the reconstruction quality, we compare the activations of the reconstructed and the original input using L1 loss and cosine similarity, averaged over \num{300} or \num{500} images. 

\begin{figure}[htb]
%\vskip 0.2in
\begin{center}
\centerline{\includegraphics[width=0.6\columnwidth]{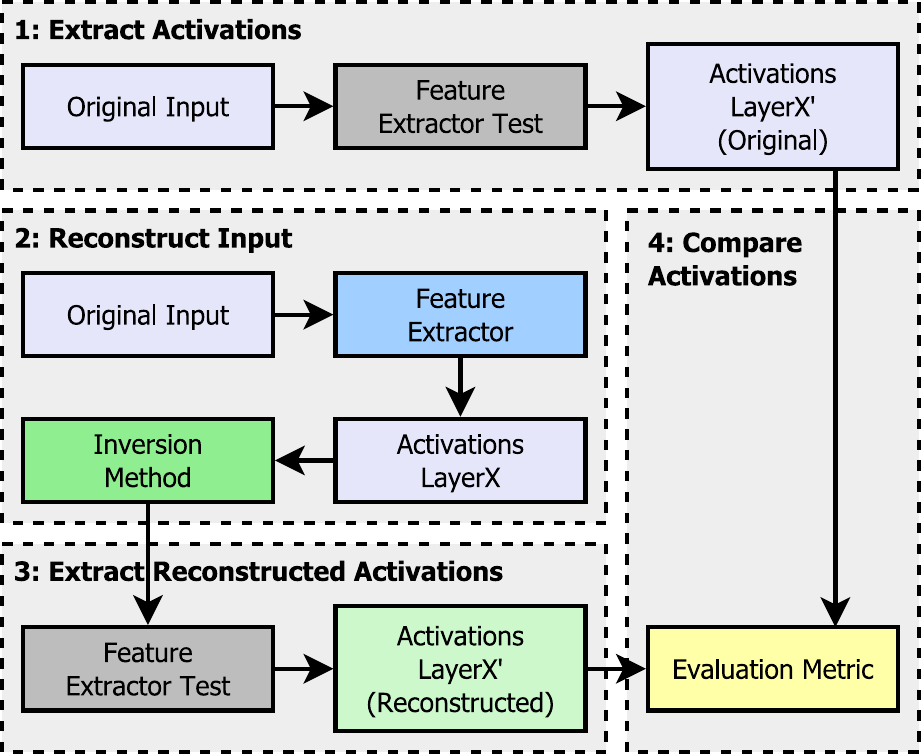}}
\caption{Comparing inversion methods at test time.}
\label{fig:inversion-methods}
\end{center}
\vskip -0.2in
\end{figure}

With this, we can investigate for a given layer in the feature extractor, which layer in the GAN reconstructs it best, i.e., which layer in the GAN is most compatible with it.
%

%To evaluate how good our GAN-based inversion method is, we compare it to gradient descent with regularization and gradient descent without regularization.
%
Similar to the training, we do not compare the original input and the result of the inversion method in the input layer but in a hidden layer of a second feature extractor.
The motivation for this is that we want to know how accurately the features extracted by LayerX were reconstructed, not whether they also match low-level features that LayerX has already discarded.
And the reason for using a second network for the comparison is to reduce overfitting effects, especially with gradient descent without regularization (see also \cref{convolutional downsampling}). %TODO rewrite differently

The whole process is shown in \cref{fig:inversion-methods}.
First, we extract the activations $\alpha^x$ from the original input from the test network at LayerX'.
Second, we generate a reconstruction of the original input from the activations at LayerX from the network we wish to invert, using either gradient descent or our GAN method for the reconstruction process.
Third, we pass the reconstruction into the second network (test network) and extract its activation $\alpha^x{}'$ at LayerX'.
Lastly, in step 4 we compare both activations $\alpha^x$ and $\alpha^x{}'$ from the test network using two different evaluation metrics (see \cref{sec:metrics}).

\subsection{Gradient Descent}

\cref{alg:invert} describes how we perform gradient descent, as proposed by \cite{olah2017feature}. % with the FFT DEC and PLAIN methods.
%
%For the PLAIN method, both the functions $g$ and $h$ in (*) are the identity, and $\sigma$ is the sigmoid function (to keep the values of the inversion in the 0 to 1 range).
When using no regularization, both the functions $g$ and $h$ in (*) are the identity function, and $\sigma$ is the sigmoid function to keep the values of the reconstructed image in a 0 to 1 range.
%
%For the FFT DEC method
When using regularization, $h$ is an inverse Fourier transform followed by correlating colors so that $z'$ is in color de-correlated Fourier space. In addition, $g$ is one-pixel jittering to reduce noise in the inversion. With one-pixel jittering, we first pad the input by 2 pixels on each side and then randomly move it up to one pixel horizontally and vertically.
We perform $n=512$ forward-backward passes through the feature extractor and use the Adam optimizer from \cite{kingma_adam_2015} with a learning rate of 0.05.

\begin{algorithm}[htb]
	\caption{Gradient Descent inversion from LayerX}
	\label{alg:invert}
	
	%\begin{spacing}{1.75}	
		\begin{algorithmic}
            \STATE {\bfseries Input:} data $x$, network from layers 0 to X $F_X$, L1 loss $L$
			\STATE $y \gets F_X(x)$
			\STATE $z'_0 \sim \mathcal{N}(0, 0.01)$
			%\STATE optimizer watches $z'_0$
			\FOR{$i \gets 0$ to $n$} 
    			\STATE $\text{loss} \gets \nabla_{z'_i} [L(F_X(g(\sigma (h(z'_i)))), y)]$ (*)
    			\STATE backpropagate loss
    			%\STATE optimizer does update step on $z'_i$
    			\STATE update $z'_i$
                \ENDFOR	 	
		\end{algorithmic}
	%\end{spacing}
\end{algorithm}

\subsection{Evaluation Metrics}

\label{sec:metrics}

Let $x \in I$ be the original input image and $F_X$ be the feature extractor from layers 0 to LayerX, i.e., the layer from which we want to reconstruct the activations. Further, let $L_X$ be the latent space of LayerX, $f: L_X \rightarrow I$ an inversion method (steps 2 - 3 in \cref{fig:overview} or gradient descent), and $F_X'$ the test network from layers 0 to LayerX'.

We compare the activations $\alpha^x{}' = F_X'(f(F_X(x)))$ and $\alpha^x = F_X'(x)$ by computing a similarity metric $d(\alpha^x{}', \alpha^x)$ with $d: L_X' \times L_X' \rightarrow \mathbb{R}$.
For $d$, we use the following two evaluation metrics:
\begin{itemize}
    \item Cosine similarity: Let $C$ be a matrix of shape (H, W) with \[C_{ij} = \frac{\alpha^x{}'[:, i, j] \cdot \alpha^x{}[:, i, j]}{\text{max}(\vert \vert \alpha^x{}'[:, i, j] \vert \vert _2 \cdot \vert \vert \alpha^x{}[:, i, j] \vert \vert _2, \epsilon)}.\] 
    Then $d(\alpha^x{}', \alpha^x) = \text{mean}(C)$. This means we first compute pixelwise cosine similarity, followed by computing the mean of the resulting matrix.
    \item L1 loss: $d(\alpha_x', \alpha_x) = \text{mean}( \vert \alpha_x' - \alpha_x \vert$)
    %\item Cosine similarity gram matrices: We first compute the gram matrices for $\alpha_x'$ and $\alpha_x$. Let $G(\alpha)$ be the gram matrix for $\alpha$, then $G(\alpha)_{ij} = \text{dot}(\text{vec}(\alpha[i]), \text{vec}(\alpha[j]))$ with $\alpha[i]$ being the i-th channel of $\alpha$ and the vec function flattening it into a vector ($\alpha[i]$ is a matrix of shape (H, W) and vec flattens it to a vector of shape $(H \cdot W)$. Then: \\
    %\[
    %d(\alpha_x', \alpha_x) = \text{sum} \left( \frac{G(\alpha_x') \cdot G(\alpha_x)}{\text{max}(\vert \vert G(\alpha_x') \vert \vert _2 \cdot \vert \vert G(\alpha_x) \vert \vert _2, \epsilon)} \right) \]
\end{itemize}
Both cosine similarity and L1 loss work pixelwise. They measure whether concepts have been correctly reconstructed and also whether they have been reconstructed at the correct position.
%
%The similarity between gram matrices can be used for style transfer \cite{styletransfer}. Therefore, with cosine similarity between gram matrices, we measure how well the concepts present in the original image have been reconstructed without depending on whether they have been correctly reconstructed spatially.
%

\subsection{Convolutional Downsampling}
\label{convolutional downsampling}

\begin{figure}[tb]
%\vskip 0.2in
\begin{center}
\centerline{\includegraphics[width=0.5\columnwidth]{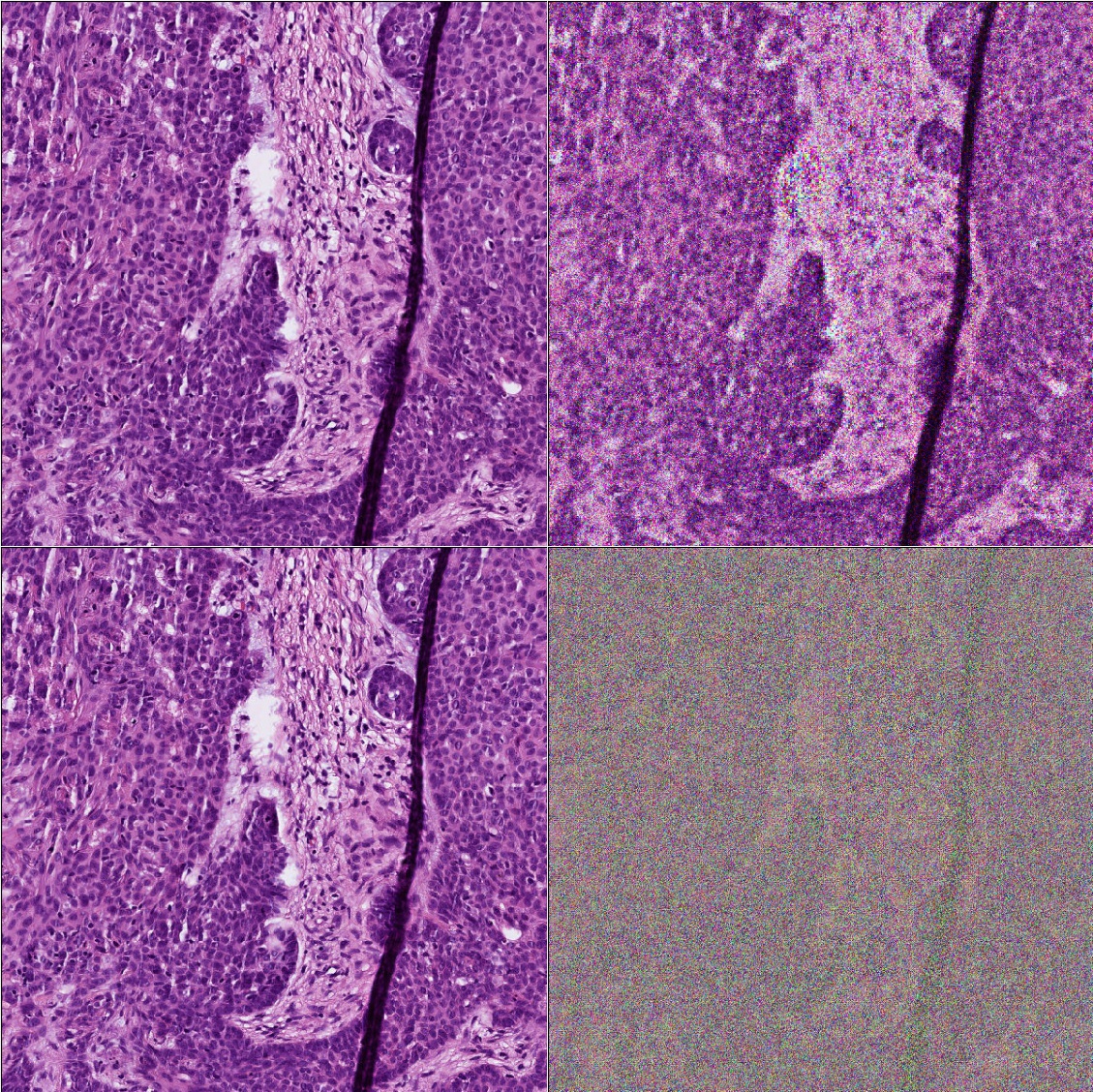}}
\caption{Comparison between inverting a ResNet34 and ResNet34 using bilinear downsampling from Layer2. The top line shows the result for a ResNet34 using bilinear downsampling and the bottom line for regular ResNet34. Left is both times the real input image, and on the right is the resulting inversion, generated using gradient descent without any regularization. Both networks are randomly initialized, and they are untrained.}
\label{fig:digipath-bilinear-conv}
\end{center}
\vskip -0.2in
\end{figure}

Strided convolutions tend to introduce checkerboard noise \citep{odena2016deconvolution}, and strided convolutional downsampling tends to introduce noise in the gradient \citep{olah2017feature}.
In the extreme setting of a 1x1 convolution with stride 2, which both the ResNet50 and ResNet34 we utilize have, only one-quarter of the input pixels would be used based on a fixed grid pattern, and the other three-quarters would be discarded.
Then only this one-quarter of pixels would receive any gradient in the backward pass, which introduces noise when reconstructing the input with gradient descent, as shown in \cref{fig:digipath-bilinear-conv}.
%

%Since the network utilizes input pixels more or less based on a fixed pattern, the noise can be reduced by jittering the input image \cite{olah2017feature}.
%
This behavior is also reflected in our evaluations: Suppose both the network we invert with gradient descent and the testing network, where we compare the reconstruction to the original image, have convolutional downsampling. In that case, the measured quality is relatively higher compared to when the testing network does not have convolutional downsampling. We additionally use VGG, which has max pooling downsampling for the AFHQ wild and ImageNet1K data, and a ResNet34 with bilinear downsampling for the Digipath data.

\section{Experiments}
\label{sec:eval}

\subsection{Setup}

In this section, we describe the datasets, models, and hardware we use in our experiments\footnote{The code is available at: https://github.com/herdtr/gan-stitching}.
%
%The code is available at: https://github.com/herdtr/gan-stitching
\subsubsection{Datasets}

We conduct our experiments on three different datasets, once on tissue scans in digital pathology data (Digipath), on Animal Faces-HQ (AFHQ) containing wild animal faces \citep{choi2020starganv2}, and on ImageNet1K \citep{deng_imagenet_2009}.
The images in the Digipath data have a spatial size of 600x600, whereas the AFHQ wild images have a size of 512x512. Here, we downsample them from 512x512 to 224x224. 
For the ImageNet1K images, we downsample and crop them to 224x224.

An in-house dataset and models that are not publicly available are used for our experiments on digital pathology data, whereas both the AFHQ and ImageNet1K datasets are available on Kaggle.
\cref{tab:datasets} shows the number of training and validation samples for the datasets we use.
For the ImageNet1K validation data, we used every hundredth image of the whole validation data of \num{50000} images resulting in \num{500} validation images.

\begin{table}[htb]
    \caption{Train- and validation set sizes for the three datasets Digipath, AFHQ wild, and ImageNet1K.}
    \label{tab:datasets}
    %\vskip 0.15in
    \begin{center}
    \begin{small}
    \begin{sc}
    \vskip -0.2in
    %\resizebox{.9\textwidth}{!}{
    \begin{tabular}{l|cccc}  %C{2cm} center column and force at 2cm
    \toprule
     & Digipath & AFHQ & Imagenet1K\\
    \midrule
    Training Size   & \num{156680} & \num{4738} & \num{1281167} \\
    Validation Size & \num{300} & \num{500} & \num{500} \\
    Used Image Size &  \num{600}x\num{600} &  \num{224}x\num{224} &  \num{224}x\num{224} \\ 
    \bottomrule
    \end{tabular}
    %}
    \end{sc}
    \end{small}
    \end{center}
    \vskip -0.1in
\end{table}

\subsubsection{Hardware and Software}

For all our experiments, we use a Linux server with 4 Nvidia RTX A6000 GPUs.
On the software side, we use PyTorch \citep{paszke2019} version 1.11.0 and Torchvision version 0.12.0.

\subsubsection{Training}

All networks in our experiments are running in eval mode.
We only train a 1x1 convolution to map between two layers. The weights of the feature extractor and the GAN generator are kept frozen.
We use Adam with a learning rate of 0.01 as an optimizer and train with a batch size of \num{8}.

\subsection{Results on the AFHQ and ImageNet1K Datasets}
In this section, we describe our results from the experiments on the AFHQ wild and ImageNet1K data.
We stitch a ResNet50 from Torchvision models trained on ImageNet, once with a StyleGAN2 trained on AFHQ wild data at a resolution of 512x512, and once with a BigGAN trained on ImageNet at a resolution of 128x128. In both cases, we scale the output of the GAN generator to 224x224 and compare the results to gradient descent with and without regularization.
As test networks, we use ResNet34 and VGG19 from Torchvision models, both trained on ImageNet.
The stitching is performed at \num{4} points of the ResNet50, namely at the output of Layer1 to Layer4.
When stitching on the AFHQ wild data, we train for \num{30} epochs, i.e., \num{142140} images in total. For the stitching on the ImageNet1K data, we train for a single epoch that involves \num{1281167} images.
%

% maybe also include accuracy for normal imgs and IoU with digipath?
\begin{table*}[htb]
\caption{Test results for the AFHQ wild data. The best results are highlighted in bold. The evaluation time is the total time needed for all \num{500} validation images.}
\label{tab:afhq-metrics}
\begin{center}
\begin{small}
\begin{sc}
\vskip -0.2in
\resizebox{1.0\textwidth}{!}{
\begin{tabular}{l|l|ccccr}  %C{2cm} center column and force at 2cm
\toprule
\multirow{3}{*}{layer} & \multirow{3}{*}{Method} &  \multicolumn{5}{c}{\textbf{AFHQ}} \\
& & Cosine Similarity & L1 Loss & Cosine Similarity VGG & L1 Loss VGG & Evaluation Time \\
\midrule
\multirow{5}{*}{Layer1}
& biggan    & 0.930 $\pm$ 0.011& 0.195 $\pm$ 0.024 & 0.733 $\pm$ 0.040 & 0.071 $\pm$ 0.009 & \textbf{4s} \\
& stylegan2    & 0.954 $\pm$ 0.008& 0.162 $\pm$ 0.019 & \textbf{0.801} $\pm$ 0.029 & \textbf{0.061} $\pm$ 0.007 & 6s \\
& stylegan2$_I$    & 0.951 $\pm$ 0.008& 0.167 $\pm$ 0.019 & 0.799 $\pm$ 0.008 & 0.062 $\pm$ 0.008 & 6s \\

& stylegan2$_{DC}$   & 0.734 $\pm$ 0.019& 0.422 $\pm$ 0.031 & 0.413 $\pm$ 0.036 & 0.119 $\pm$ 0.009 & 6s \\

& gd with reg & 0.933 $\pm$ 0.010&  0.196 $\pm$ 0.023 & 0.705 $\pm$ 0.032 & 0.078 $\pm$ 0.009 & 251s \\ %259s \\
& gd no reg   & \textbf{0.974} $\pm$ 0.007&  \textbf{0.117} $\pm$ 0.020 & 0.722 $\pm$ 0.033 & 0.074 $\pm$ 0.006 & 238s \\ %246s \\
\midrule
\multirow{5}{*}{Layer2}
& biggan    & 0.835 $\pm$ 0.017&  0.158 $\pm$ 0.013 & 0.614 $\pm$ 0.032  & 0.054 $\pm$ 0.004 & \textbf{3s} \\
& stylegan2    & 0.878 $\pm$ 0.015&  0.135 $\pm$ 0.011 & \textbf{0.716} $\pm$ 0.024 & 0.046 $\pm$ 0.003 & 6s \\
& stylegan2$_I$    & 0.871 $\pm$ 0.015&  0.139 $\pm$ 0.011 & 0.707 $\pm$ 0.024 & 0.047 $\pm$ 0.003 & 6s \\
& stylegan2$_{DC}$    & 0.540 $\pm$ 0.018&  0.288 $\pm$ 0.014 & 0.264 $\pm$ 0.015 & 0.083 $\pm$ 0.004 & 6s \\
& gd with reg & 0.905 $\pm$ 0.009&  0.119 $\pm$ 0.008 & 0.711 $\pm$ 0.023 & \textbf{0.045} $\pm$ 0.003 & 395s \\ %401s \\
& gd no reg    & \textbf{0.931} $\pm$ 0.013&  \textbf{0.098} $\pm$ 0.011 & 0.670 $\pm$ 0.024 & 0.048 $\pm$ 0.003 & 380s \\ % 386s \\
\midrule
\multirow{6}{*}{Layer3}
& biggan    & 0.316 $\pm$ 0.018&  0.137 $\pm$ 0.005 & 0.180 $\pm$ 0.018 & 0.031 $\pm$ 0.001 & \textbf{3s} \\
& biggan+1    & 0.724 $\pm$ 0.038&  0.083 $\pm$ 0.009 & 0.486 $\pm$ 0.042 & 0.026 $\pm$ 0.002 & \textbf{3s} \\
& stylegan2    & 0.778 $\pm$ 0.032&  0.070 $\pm$ 0.007 & 0.579 $\pm$ 0.039 & 0.023 $\pm$ 0.002 & 6s \\
& stylegan2$_I$    & 0.741 $\pm$ 0.031&  0.077 $\pm$ 0.007 & 0.523 $\pm$ 0.033 & 0.025 $\pm$ 0.002 & 6s \\
& stylegan2$_{DC}$    & 0.309 $\pm$ 0.020&  0.143 $\pm$ 0.006 & 0.162 $\pm$ 0.016 & 0.032 $\pm$ 0.002 & 6s \\
& gd with reg & \textbf{0.840} $\pm$ 0.023 &  \textbf{0.060} $\pm$ 0.006 & \textbf{0.651} $\pm$ 0.036 & \textbf{0.021} $\pm$ 0.002 & 530s \\ %536s \\
& gd no reg    & 0.762 $\pm$ 0.036& 0.074 $\pm$ 0.008 & 0.492 $\pm$ 0.036 & 0.025 $\pm$ 0.002 & 515s \\ %521s \\
\midrule
\multirow{3}{*}{Layer4} 
& stylegan2    & 0.527 $\pm$ 0.117& 0.868 $\pm$ 0.108 & 0.287 $\pm$ 0.095 & 0.128 $\pm$ 0.013 & \textbf{6s} \\ %TODO train gan longer here
& gd with reg & \textbf{0.697} $\pm$ 0.054&  \textbf{0.763} $\pm$ 0.089 & \textbf{0.428} $\pm$ 0.061 & \textbf{0.121} $\pm$ 0.012 & 589s \\ %595s \\
& gd no reg    & 0.341 $\pm$ 0.029&  0.945 $\pm$ 0.060 & 0.153 $\pm$ 0.027 & 0.122 $\pm$ 0.013 & 574s \\ % 580s \\
\bottomrule
\end{tabular}
}
\end{sc}
\end{small}
\end{center}
\vskip -0.1in
\end{table*}

\begin{table*}[tb]
\caption{Test results for the ImageNet1K data. The best results are highlighted in bold. The evaluation time is the total time needed for all \num{500} validation images.}
\label{tab:imagenet-metrics}
%\vskip 0.15in
\begin{center}
\begin{small}
\begin{sc}
\vskip -0.2in
\resizebox{1.0\textwidth}{!}{
\begin{tabular}{l|l|ccccr}  %C{2cm} center column and force at 2cm
\toprule
\multirow{3}{*}{layer} & \multirow{3}{*}{Method} &  \multicolumn{5}{c}{\textbf{ImageNet1K}} \\
& & Cosine Similarity & L1 Loss & Cosine Similarity VGG & L1 Loss VGG & Evaluation Time \\
\midrule
\multirow{5}{*}{Layer1}
& biggan    & 0.907 $\pm$ 0.028& 0.230 $\pm$ 0.049 & 0.696 $\pm$ 0.087 & 0.080 $\pm$ 0.018 & \textbf{3s} \\
& stylegan2    & 0.926 $\pm$ 0.023&  0.209 $\pm$ 0.043 & 0.768 $\pm$ 0.059 & 0.070 $\pm$ 0.014 & 6s   \\
& stylegan2$_I$    & 0.931 $\pm$ 0.022 &  0.202 $\pm$ 0.042 & \textbf{0.775} $\pm$ 0.059 & \textbf{0.068} $\pm$ 0.015 & 6s \\
& gd with reg & 0.911 $\pm$ 0.021 &  0.235 $\pm$ 0.043 & 0.710 $\pm$ 0.046 & 0.081 $\pm$ 0.013 & 251s \\ %261s \\
& gd no reg    & \textbf{0.955} $\pm$ 0.014 & \textbf{0.159} $\pm$ 0.034 & 0.707 $\pm$ 0.056 & 0.079 $\pm$ 0.010 & 238s \\ %246s \\
\midrule
\multirow{5}{*}{Layer2}
& biggan    & 0.818 $\pm$ 0.030&  0.163 $\pm$ 0.022 & 0.591 $\pm$ 0.055 & 0.054 $\pm$ 0.007 & \textbf{3s} \\
& stylegan2    & 0.833 $\pm$ 0.026 & 0.157 $\pm$ 0.018 & 0.645 $\pm$ 0.043 & 0.051 $\pm$ 0.005 & 6s \\
& stylegan2$_I$    & 0.848 $\pm$ 0.024 &  0.150 $\pm$ 0.018 & 0.666 $\pm$ 0.047 & 0.050 $\pm$ 0.006 & 6s \\
& gd with reg & 0.883 $\pm$ 0.015 &  0.131 $\pm$ 0.012 & \textbf{0.675} $\pm$ 0.032 & \textbf{0.046} $\pm$ 0.004 & 395s \\ %401s \\
& gd no reg    & \textbf{0.909} $\pm$ 0.020 &  \textbf{0.112} $\pm$ 0.015 &  0.646 $\pm$ 0.034 & 0.048 $\pm$ 0.004 & 380s \\ %386s \\
\midrule
\multirow{8}{*}{Layer3}
& biggan    & 0.376 $\pm$ 0.053 &  0.137 $\pm$ 0.015 & 0.213 $\pm$ 0.050 & 0.030 $\pm$ 0.003 & \textbf{3s} \\
& biggan+1    & 0.676 $\pm$ 0.044 &  0.099 $\pm$ 0.017 & 0.440 $\pm$ 0.053 & 0.026 $\pm$ 0.003 & \textbf{3s} \\
& stylegan2    & 0.557 $\pm$ 0.070 & 0.116 $\pm$ 0.021 & 0.361 $\pm$ 0.061 & 0.029 $\pm$ 0.003 & 6s \\
& stylegan2+1    & 0.583 $\pm$ 0.062 & 0.112 $\pm$ 0.020 & 0.374 $\pm$ 0.056 & 0.028 $\pm$ 0.003 & 6s \\
& stylegan2$_I$    & 0.634 $\pm$ 0.048 & 0.103 $\pm$ 0.018 & 0.421 $\pm$ 0.050 & 0.027 $\pm$ 0.003 & 6s \\
& stylegan2$_I$+1    & 0.669 $\pm$ 0.044 & 0.097 $\pm$ 0.017 & 0.449 $\pm$ 0.047 & 0.026 $\pm$ 0.003 & 6s \\
& gd with reg & \textbf{0.782} $\pm$ 0.041 &  \textbf{0.079} $\pm$ 0.013 & \textbf{0.574} $\pm$ 0.053 & \textbf{0.022} $\pm$ 0.003 & 530s \\ %535s \\
& gd no reg    & 0.725 $\pm$ 0.050 & 0.089 $\pm$ 0.015 & 0.468 $\pm$ 0.050 & 0.025 $\pm$ 0.003 & 515s \\ %520s \\
%\midrule
%\multirow{3}{*}{Layer4} 
%& stylegan2    & - $\pm$ 0.117& 0.518$\pm$ 0.179 & 0.868 $\pm$ 0.108 & \textbf{7s} \\ %TODO train gan longer here
%& fft dec & 0.570 $\pm$ 0.098 & 0.506 $\pm$ 0.135 & 0.978 $\pm$ 0.146 & 594s \\
%& plain    & 0.369 $\pm$ 0.067 & 0.215 $\pm$ 0.076 & 1.046 $\pm$ 0.101 & 577s \\
\bottomrule
\end{tabular}
}
\end{sc}
\end{small}
\end{center}
\vskip -0.1in
\end{table*}

\cref{tab:afhq-metrics} shows the results of different metrics when inverting a ResNet50 classifier trained on ImageNet data. It compares different inversion methods and measures their run time for the 500 validation images of the AFHQ data. The "Evaluation Time" column shows the accumulated time over the \num{500} images.
\cref{tab:imagenet-metrics} evaluates the different inversion methods on the 500 validation images of the ImageNet1K data.
In \cref{tab:afhq-metrics}, we report the metrics on the validation data for the epoch which had the highest cosine similarity on the validation data and in \cref{tab:imagenet-metrics} we report the metrics after one epoch of training.

We do the reconstruction at the output of Layer1 to Layer4 of ResNet50, shown in the ''layer'' column, each time with six different methods: 
\begin{enumerate}
    \item BigGAN where the 1x1 stitching convolution is trained on the ImageNet1K data
    \item StyleGAN2 where the 1x1 stitching convolution is trained on the AFHQ wild data 
    \item StyleGAN2 where the 1x1 stitching convolution is trained on the ImageNet1K data, denoted as $\text{StyleGAN2}_I$
    \item StyleGAN2 without the 1x1 stitching convolution (i.e. directly copying the intermediate output of the feature extractor into the GAN), denoted as $\text{StyleGAN2}_{DC}$    
    \item By gradient descent with regularization (GD with reg)
    \item By gradient descent without any regularization (GD no reg).
\end{enumerate}

\begin{figure*}[tb]
%\vskip 0.2in
\begin{center}
\centerline{\includegraphics[width=\columnwidth]{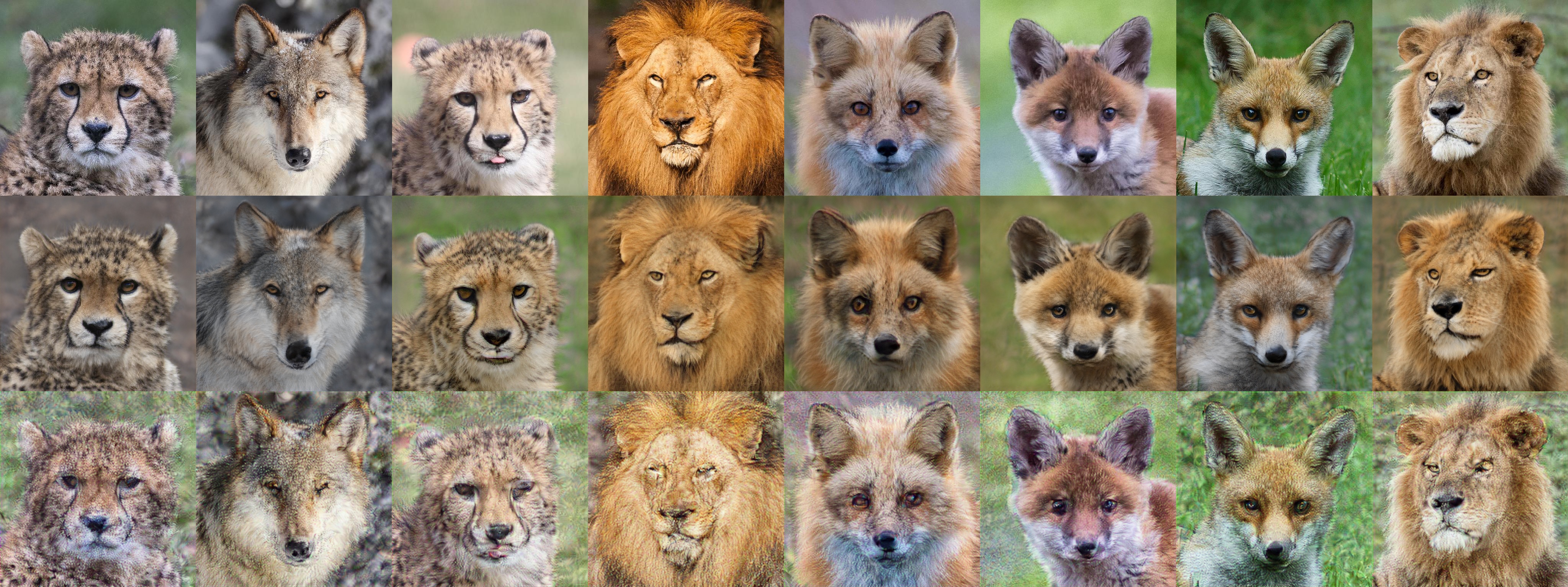}}
\caption{Input reconstruction from activations at the output of Layer3 of ResNet50 for \num{8} test samples where the GAN method had the highest cosine similarity. The top row shows the original dataset images, the middle line shows their reconstructions from StyleGAN2, and the bottom line shows their reconstructions using gradient descent with regularization.}
\label{fig:sample-imgs-Layer3}
\end{center}
\vskip -0.2in
\end{figure*}

For the GAN methods, we need not only a starting layer in ResNet50 but also a layer in the GAN that we want to stitch into.
Unless noted otherwise, we choose the first convolution of a layer that is the same number of samplings away from the output as the starting layer in ResNet50 is away from the input.
Exemplary, for StyleGAN2 that means we stitch
\begin{itemize}
    \item ResNet50 Layer1 $\rightarrow$ StyleGAN2 b128.conv0 (Layer1 is two downsampling steps away from the input of ResNet50, and b128.conv0 is two upsampling steps away from the output of StyleGAN2)
    \item ResNet50 Layer2 $\rightarrow$ StyleGAN2 b64.conv0
    \item ResNet50 Layer3 $\rightarrow$ StyleGAN2 b32.conv0
    \item ResNet50 Layer4 $\rightarrow$ StyleGAN2 b16.conv0
\end{itemize}
The notation of the form "+1", as used in \cref{tab:imagenet-metrics} for Layer3, refers to stitching into a layer that is one level shallower.
In \cref{sec:Different End Layer}, we investigate different combinations of stitching layers.
Cosine similarity and L1 loss are evaluated for each of the \num{500} validation images. The tables show the mean $\pm$ standard deviation. 

The GAN methods take considerably less time than the gradient descent methods, as shown in the ''Evaluation Time'' column.
For measuring the time of the GAN methods, we let it run over the dataset \num{10} times and take the mean run time.
The reported evaluation times do not include the time for computing the evaluation metrics, only for computing the reconstructions.
For example, for Layer3, the GAN methods take between 3 and 6 seconds, whereas gradient descent takes more than 500 seconds.
%
% TODO: replace runtime improvements by new values
% TODO: replace gradient descent runtimes by new values that do not include evaluation metrics, but only reconstruction (like with the gan)
In our experiments, we observed an average runtime improvement with StyleGAN2 of 73x with a minimum improvement of 39x and a maximum improvement of 101x compared to gradient descent without regularization.
Similar improvements can also be observed compared to gradient descent with regularization.
In this case, we observed an average runtime improvement of 76x with a minimum of 41x and a maximum of 103x.
Our method runs faster because it only needs one forward pass through the feature extractor and the GAN generator instead of \num{512} forward-backward passes through the feature extractor with gradient descent.

%

%Usually, the GAN method performs best for the cosine similarity between gram matrices of activations metric (except for Layer4).
%
In Layer3 and Layer4, gradient descent with regularization performs best in cosine similarity and L1 loss. In earlier layers, it is worse than gradient descent without any regularization, likely due to the effect of one-pixel jittering. %due to the effect of one-pixel jittering (without it in Layer2 we get an L1 loss for FFT DEC of 0.078 vs.\ 0.101 for PLAIN and in Layer1 0.096 vs.\ 0.120 for PLAIN, both times outperforming PLAIN).
We use the one-pixel jittering to reduce noise in the reconstruction, but it forces that even if we move the reconstruction one pixel, it should still have similar activations in the hidden layer. Combining that with the fact that 1x1 (and to some extent also 3x3) convolutions with stride 2 utilize pixels in a grid pattern likely reduces the evaluation score for the ResNet34.
For VGG, which uses max pooling for downsampling, gradient descent with regularization outperforms gradient descent without regularization from Layer2 downwards, and in Layer1 the GAN methods are best.

Directly copying the activations into the GAN (i.e. reconstructing without using the 1x1 stitching convolution) does not work, as shown in \cref{fig:direct_copy} and \cref{tab:afhq-metrics}.
This is expected, since the feature extractor and GAN where trained completely unaware of each other.

\cref{fig:sample-imgs-Layer3} shows \num{8} samples for reconstructing activation from Layer3 (with using the 1x1 stitching convolution) using StyleGAN2 and gradient descent with regularization.
We can see that the reconstruction with StyleGAN2 does not introduce noise in the image, whereas gradient descent does.
But the GAN method loses lighting and some color information.

\begin{figure*}[tb]
%\vskip 0.2in
\begin{center}
\centerline{\includegraphics[width=\columnwidth]{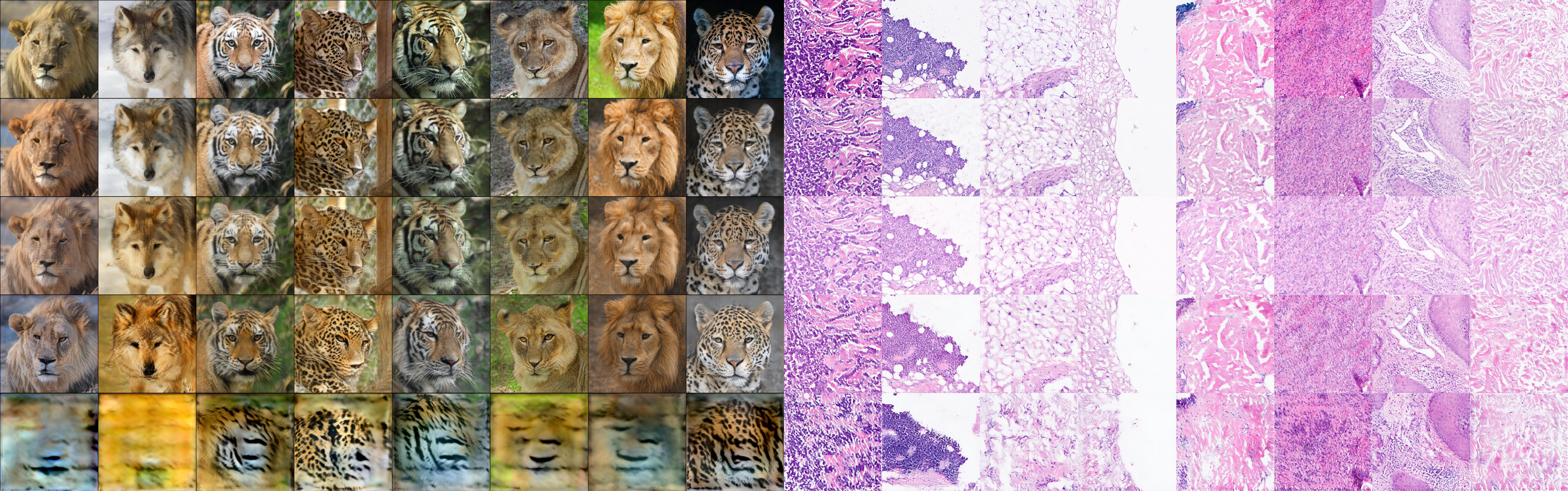}}
\caption{Comparison of the original images in the first row with our GAN-reconstructed activations down from Layer1 to Layer4 in the following rows. Left using StyleGAN2 on AFHQ data, right using a custom GAN on Digipath data.}
\label{fig:different-layers}
\end{center}
\vskip -0.2in
\end{figure*}

\cref{fig:different-layers} shows reconstructions using StyleGAN2 for Layer1 to Layer4 for the AFHQ data, and using our custom GAN for the Digipath data.
The top line contains the original images, and the following four lines show reconstructions from Layer1 to Layer4.
Each column for the AFHQ wild data uses the same noise seed (same $z$ vector for StyleGAN2).
The reconstruction works, except for Layer4 for the AFHQ wild data, where it is reduced to a fur pattern (tiger fur pattern for a tiger image and leopard fur pattern for a leopard image).
For Layer1 and Layer2, it mainly retains lower-level information, like the fur pattern of a leopard. The image also becomes slightly blurred by reconstructing it from Layer2. In contrast, in Layer3, it hallucinates a new fur pattern, and the image becomes sharper again.
%
%It is also interesting that for Layer4, the reconstructions for the AFHQ wild data are broken, whereas, for the Digipath data, they still resemble the general shape of the original input image.
%
%That may be because the StyleGAN2 was only trained on AFHQ wild data. In contrast, the ResNet50 we want to interpret was trained on ImageNet (much more variety of concepts), whereas, for the Digipath data, both the GAN and the segmentation network were trained on primarily the same data.
%

\subsection{Digipath Dataset Results}

\begin{table*}[htb]
\caption{Test results for Digipath data. The best results are highlighted in bold. The evaluation time is the total time needed for all \num{300} validation images.}
\label{tab:digipath-metrics}
%\vskip 0.15in
\begin{center}
\begin{small}
\begin{sc}
\vskip -0.2in
\resizebox{.7\textwidth}{!}{
\begin{tabular}{l|l|ccr}  %C{2cm} center column and force at 2cm
\toprule
\multirow{3}{*}{layer} & \multirow{3}{*}{Method} &  \multicolumn{3}{c}{\textbf{Digipath}} \\
& & Cosine Similarity& L1 Loss & Evaluation Time \\
\midrule
\multirow{3}{*}{Layer1}
& gan    & \textbf{0.896} $\pm$ 0.018 & \textbf{0.601} $\pm$ 0.100 & \textbf{14s} \\
& gd with reg & 0.825 $\pm$ 0.043 & 0.786 $\pm$ 0.189 & 1112s \\
& gd no reg    & 0.721 $\pm$ 0.098 & 0.948 $\pm$ 0.067 & 1044s \\
\midrule
\multirow{3}{*}{Layer2}
& gan    & \textbf{0.857} $\pm$ 0.028& \textbf{0.731} $\pm$ 0.089 & \textbf{10s} \\
& gd with reg & 0.832 $\pm$ 0.037 & 0.771 $\pm$ 0.057 & 1306s \\
& gd no reg    & 0.691 $\pm$ 0.045& 1.069 $\pm$ 0.053 & 1237s \\
\midrule
\multirow{3}{*}{Layer3}
& gan    & \textbf{0.808} $\pm$ 0.035 & \textbf{0.904} $\pm$ 0.084 & \textbf{11s} \\
& gd with reg & 0.795 $\pm$ 0.023& 0.931 $\pm$ 0.069 & 1496s \\
& gd no reg    & 0.489 $\pm$ 0.047 & 1.482 $\pm$ 0.156 & 1426s \\
\midrule
\multirow{3}{*}{Layer4} 
& gan    & \textbf{0.743} $\pm$ 0.067 & \textbf{0.788} $\pm$ 0.086 & \textbf{10s} \\
& gd with reg & 0.711 $\pm$ 0.055 & 0.842 $\pm$ 0.074 & 1584s \\
& gd no reg    & 0.317 $\pm$ 0.060 & 1.403 $\pm$ 0.141 & 1513s \\
\bottomrule
\end{tabular}
}
\end{sc}
\end{small}
\end{center}
\vskip -0.1in
\end{table*}

In this section, we describe our results for the digital pathology data.
\subsubsection{Models}

We stitch a semantic segmentation network using a ResNet34 backbone trained to segment different diseases in tissue slides in digital pathology, with a GAN generator trained unsupervised to generate tissue slides.
For the test network, we use a slightly modified ResNet34 as the backbone, where we replaced the 3x3 convolutional downsamplings with bilinear downsampling followed by a 3x3 convolution  with stride 1.
Further, we removed the 1x1 convolutional downsampling skip connections and the max pooling layer.
Those adjustments were made to reduce noise in the gradient, as visible in \cref{fig:digipath-bilinear-conv}.

\subsubsection{Results}

We train the 1x1 convolution for \num{240000} samples.
\cref{tab:digipath-metrics} shows our results evaluated on \num{300} images of the Digipath data.
The table can be read similarly to \cref{tab:afhq-metrics}.
Again, the GAN method takes considerably less time than the gradient descent methods.
We observed an average speedup of 120x with a minimum improvement of 76x and a maximum of 147x compared to gradient descent without regularization.
Moreover, we observed a similar speedup compared to gradient descent with regularization, with an average improvement of 126x, a minimum of 81x, and a maximum of 154x.

Two results are noticeably different compared to the results for the AFHQ wild data.
First, the GAN method performs best in the cosine similarity and L1 loss metrics over all layers, whereas for the AFHQ wild and ImageNet1K data, it only did for Layer1 testing in VGG.

Second, gradient descent without regularization performs considerably worse here compared to gradient descent with regularization, even in the earlier layers.
A likely reason for those two differences is that the test network we use for the Digipath data does not use convolutional downsampling but bilinear downsampling. Therefore, it is not using input pixels in the same noise pattern as the reconstruction network.
%
%Removing the max pooling layer and the 1x1 convolutional downsamplings in the skip connection and replacing the 3x3 convolutional downsamplings with bilinear downsamplings followed by a 3x3 stride 1 convolution dramatically reduces noise in inversions with gradient descent, as shown in \cref{fig:digipath-bilinear-conv}.
%
%
%1x1 stride 2 convolutions only utilize one-quarter of the input pixels in a grid pattern. Three-quarters of the input pixels remain unused. Those three-quarters will receive zero gradients, leading to a checkerboard noise pattern in the gradient.
%
%3x3 stride 2 convolutions can also utilize some pixels more than others, based on a grid pattern \cite{odena2016deconvolution}.
%
%For the AFHQ data, both ResNet50 and ResNet34 use convolutional downsamplings and share the same grid pattern.
%
%However, for the Digipath data, the test network does not use convolutional downsamplings, which means it is not assigning a magnitude of importance to pixels based on grid patterns, like the network we invert is doing.
%
%For this reason, we think that the noise patterns that gradient descent without any regularization (and to some extent also with regularization) is building into the inverted image do not transfer very well over to the test network. As a result, the gradient descent inversions perform worse compared to the GAN method.
%
Consequently, since gradient descent without any regularization is building in more noise, it is even stronger penalized in the metrics and performs considerably worse compared to the AFHQ wild and ImageNet1K data.

\subsection{Different End Layer}
\label{sec:Different End Layer}

\begin{figure*}[ht]
%\vskip 0.2in
\begin{center}
\centerline{\resizebox{\linewidth}{!}{
    \centering
    \begin{minipage}{1.15\textwidth}
        \centering
        \includegraphics{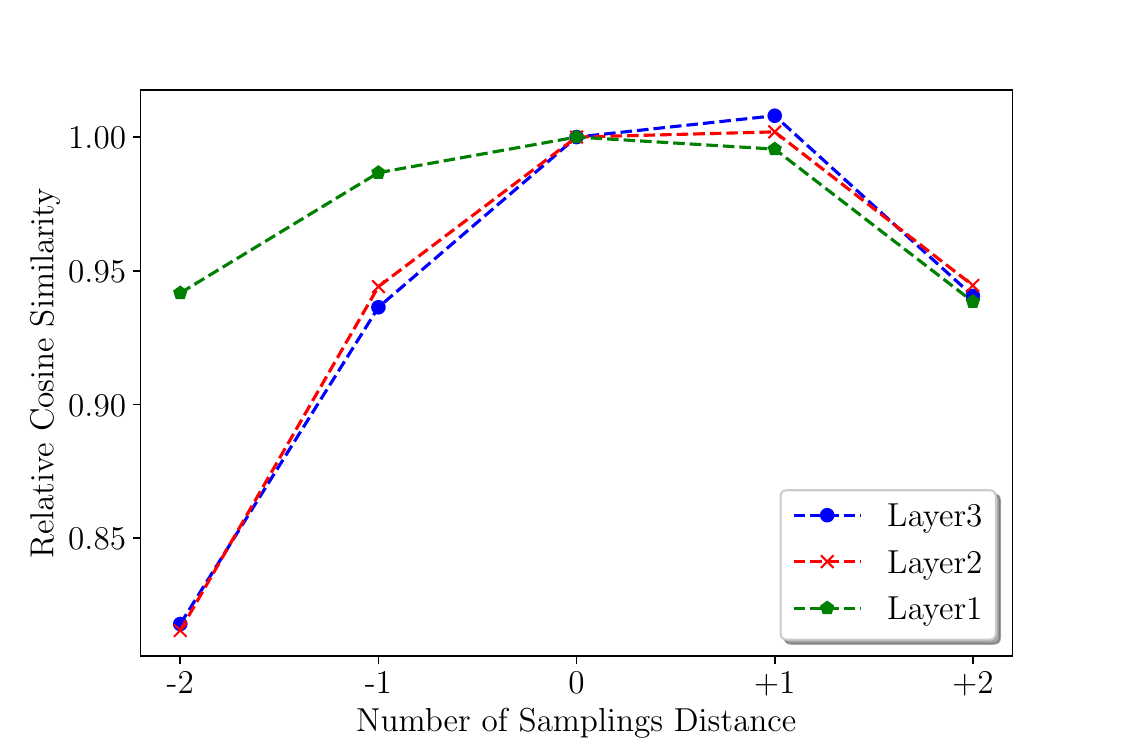}
        %\caption{$dt=0.1$}
        %\label{fig:prob1_6_2}
    \end{minipage}%
    \begin{minipage}{1.15\textwidth}
        \centering
        \includegraphics{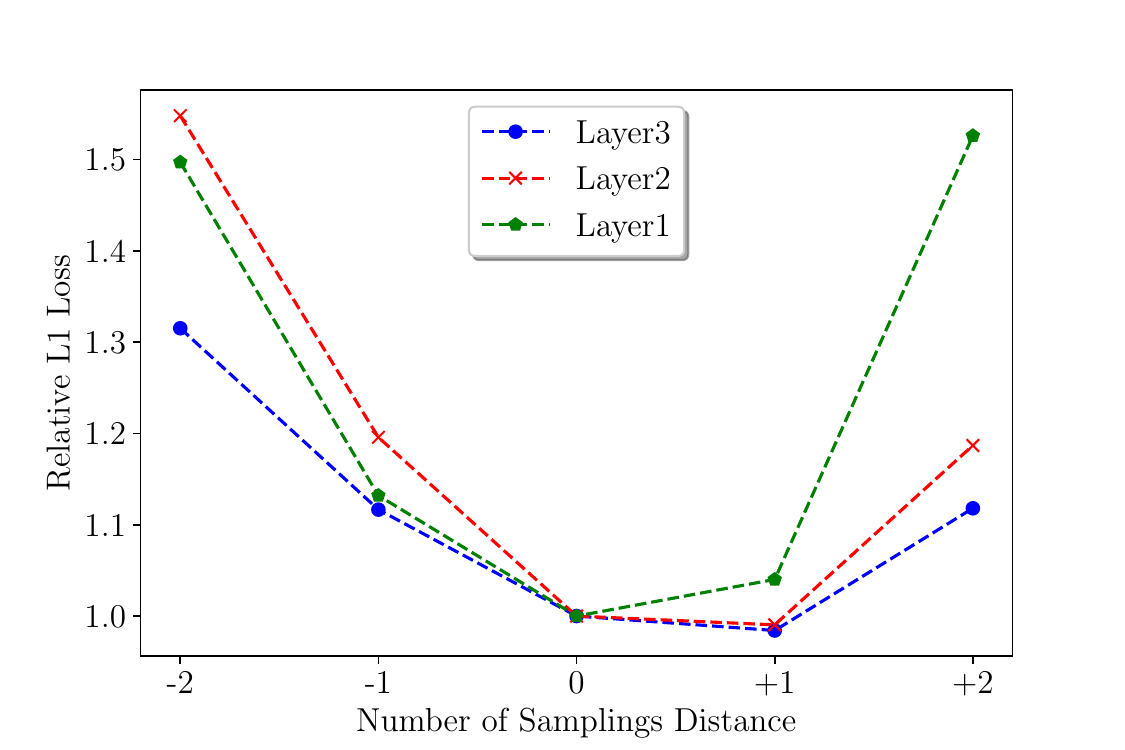}
        %\caption{$dt =$}
        %\label{fig:prob1_6_1}
    \end{minipage}
}}
\vskip -0.05in
\centerline{\resizebox{\linewidth}{!}{
    \centering
    \begin{minipage}{1.15\textwidth}
        \centering
        \includegraphics{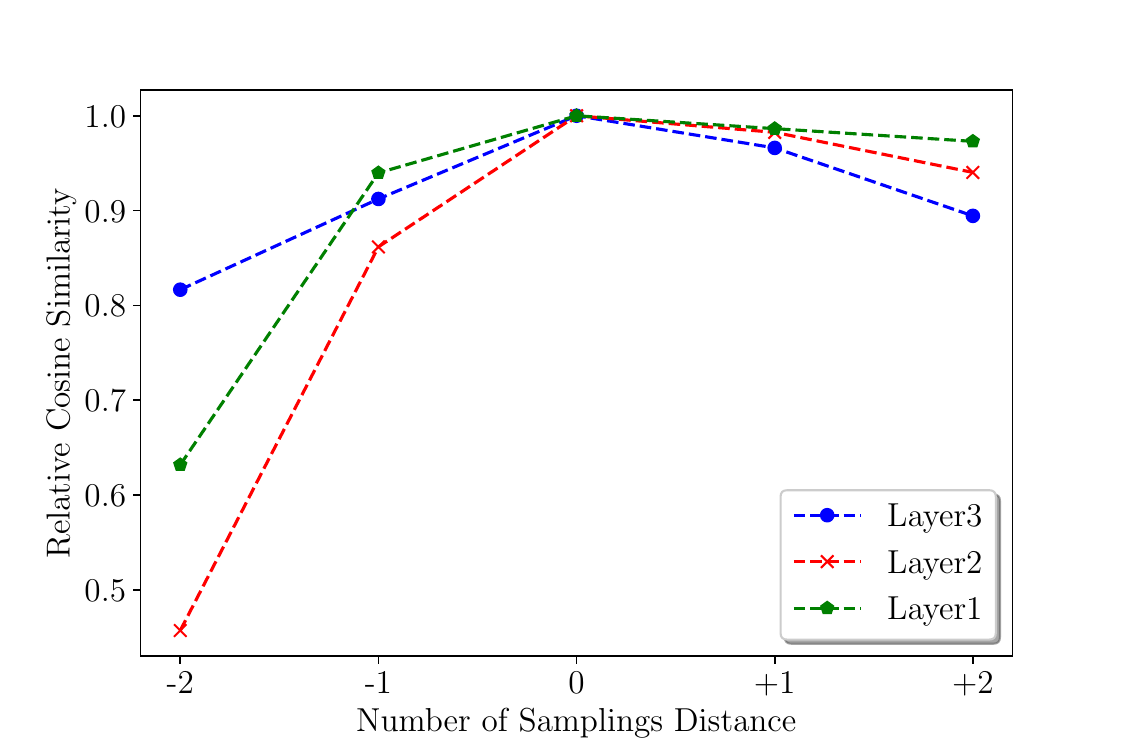}
        %\caption{$dt=0.1$}
        %\label{fig:prob1_6_2}
    \end{minipage}%
    \begin{minipage}{1.15\textwidth}
        \centering
        \includegraphics{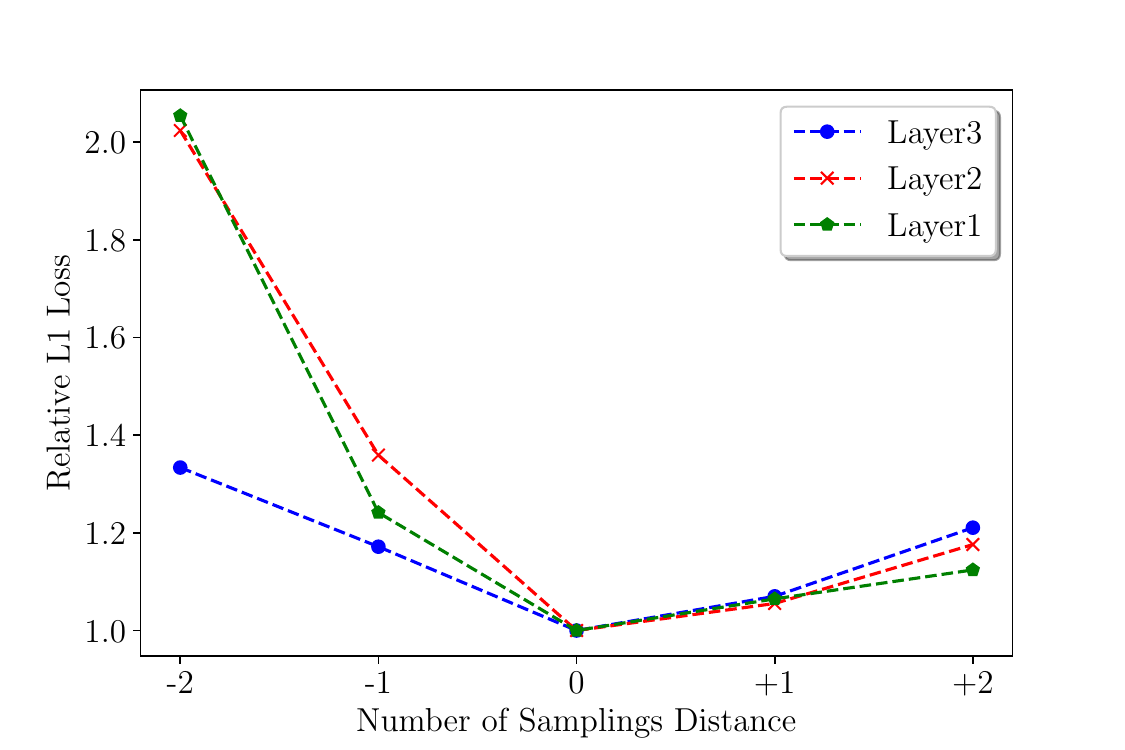}
        %\caption{$dt =$}
        %\label{fig:prob1_6_1}
    \end{minipage}
}}
\caption{Result for stitching into different target layers. The top line shows results for StyleGAN2 on AFHQ wild data, and the bottom line for our custom GAN on Digipath data.}
\label{fig:different-end-layer}
\end{center}
\vskip -0.2in
\end{figure*}

%So far, we have stitched only into one GAN layer for a given feature extractor layer.
So far, for a given layer in the feature extractor, we only stitched into one layer of the GAN.
Here, we evaluate the results of stitching from one layer of the feature extractor into different layers of the GAN to investigate which combination of layers is more compatible.
\cref{fig:different-end-layer} shows the relative cosine similarity and L1 loss of reconstructions when stitching into different layers of StyleGAN2 (top line) and the GAN generator we use for the Digipath data (bottom line).
In \cref{tab:afhq-metrics} and \cref{tab:digipath-metrics}, we stitch into a layer of the GAN generator which has the same number of samplings distance from the output, as the layer in the feature extractor we start from has from the input.
That combination would correspond to a ''Number of Samplings Distance'' of 0 in \cref{fig:different-end-layer}.
Here, -1 means we stitch into a layer in the GAN generator, which is one upsampling step further away from the output of the generator, and +1 means we stitch into a layer which is one upsampling step closer to the output of the generator.
As an example, \cref{tab:stitching-target-layer} shows the different target layers in StyleGAN2 when starting from Layer3 of ResNet50.
%

\iffalse
\begin{table}[htb]
\caption{Target layers in StyleGAN2 when starting from Layer3.}
\label{tab:stitching-target-layer}
\vskip 0.15in
\begin{center}
\begin{small}
\begin{sc}
%\resizebox{.9\textwidth}{!}{
\begin{tabular}{cr}  %C{2cm} center column and force at 2cm
\toprule
 Number of Samplings Distance & Target layer \\
\midrule
+2 & b128.conv0 \\
+1 & b64.conv0 \\
0    & b32.conv0 \\
-1 & b16.conv0 \\
-2 & b8.conv0 \\
\bottomrule
\end{tabular}
%}
\end{sc}
\end{small}
\end{center}
\vskip -0.1in
\end{table}
\fi

\begin{table}[htb]
\caption{Target layers in StyleGAN2 when starting from Layer3.}
\label{tab:stitching-target-layer}
%\vskip 0.15in
\begin{center}
\begin{small}
\begin{sc}
\vskip -0.2in
%\resizebox{.9\textwidth}{!}{
\begin{tabular}{l|ccccc}  %C{2cm} center column and force at 2cm
\toprule
Samplings Distance & +2 & +1 & 0 & -1 & -2\\
 Target layer & b128.conv0 & b64.conv0 & b32.conv0 & b16.conv0 & 8.conv0\\
\bottomrule
\end{tabular}
%}
\end{sc}
\end{small}
\end{center}
\vskip -0.1in
\end{table}

We plot the values relative to the value at 0 distance, and before plotting, we divide the value by the value at 0 distance.
For the AFHQ wild data for Layer2 and Layer3, stitching into a layer with a +1 distance is slightly better compared to stitching into 0 distance, whereas for Layer1, it is a bit worse.
For the Digipath data, it is best to stitch into a layer with 0 distance.

\subsection{Additional Discussion}

\begin{figure*}[ht]
%\vskip 0.2in
\begin{center}
\centerline{\resizebox{\linewidth}{!}{
    \centering
    \begin{minipage}{1.4\textwidth}
        \centering
        \centerline{\includegraphics[width=1.0\columnwidth]{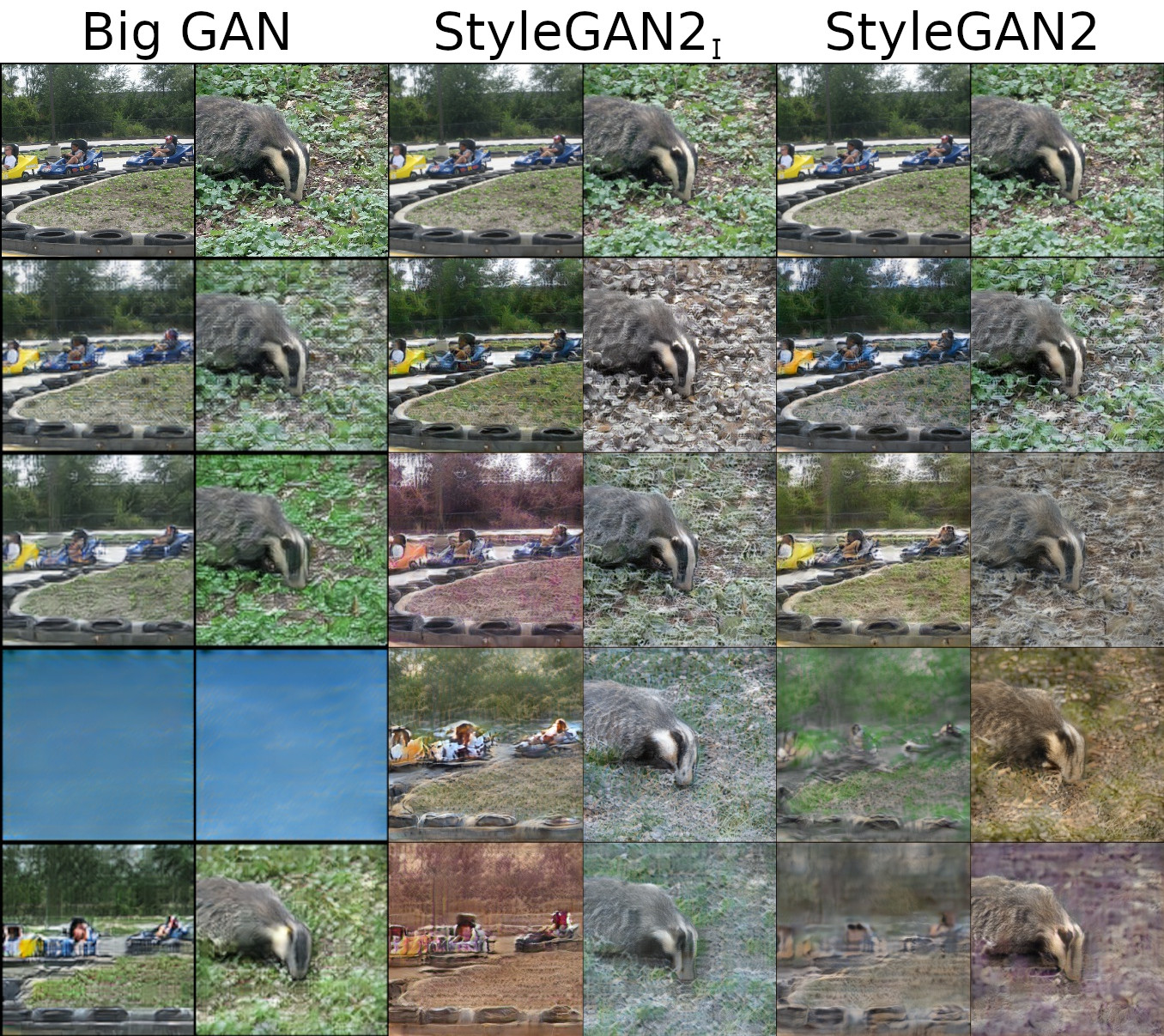}}
        \caption{Samples for ImageNet1K. The rows from top to bottom are: The original dataset image, reconstruction from Layer1, Layer2, Layer3, and Layer3 with stitching into +1.}% From left to right in columns of two: Reconstructions using BigGAN, StyleGAN2, and StyleGAN2$_I$.}
        \label{fig:imagenet-samples}
    \end{minipage}                  \, \, \, \, \, \, 
    \begin{minipage}{1.4\textwidth}
        \centering
        \centerline{\includegraphics[width=\columnwidth]{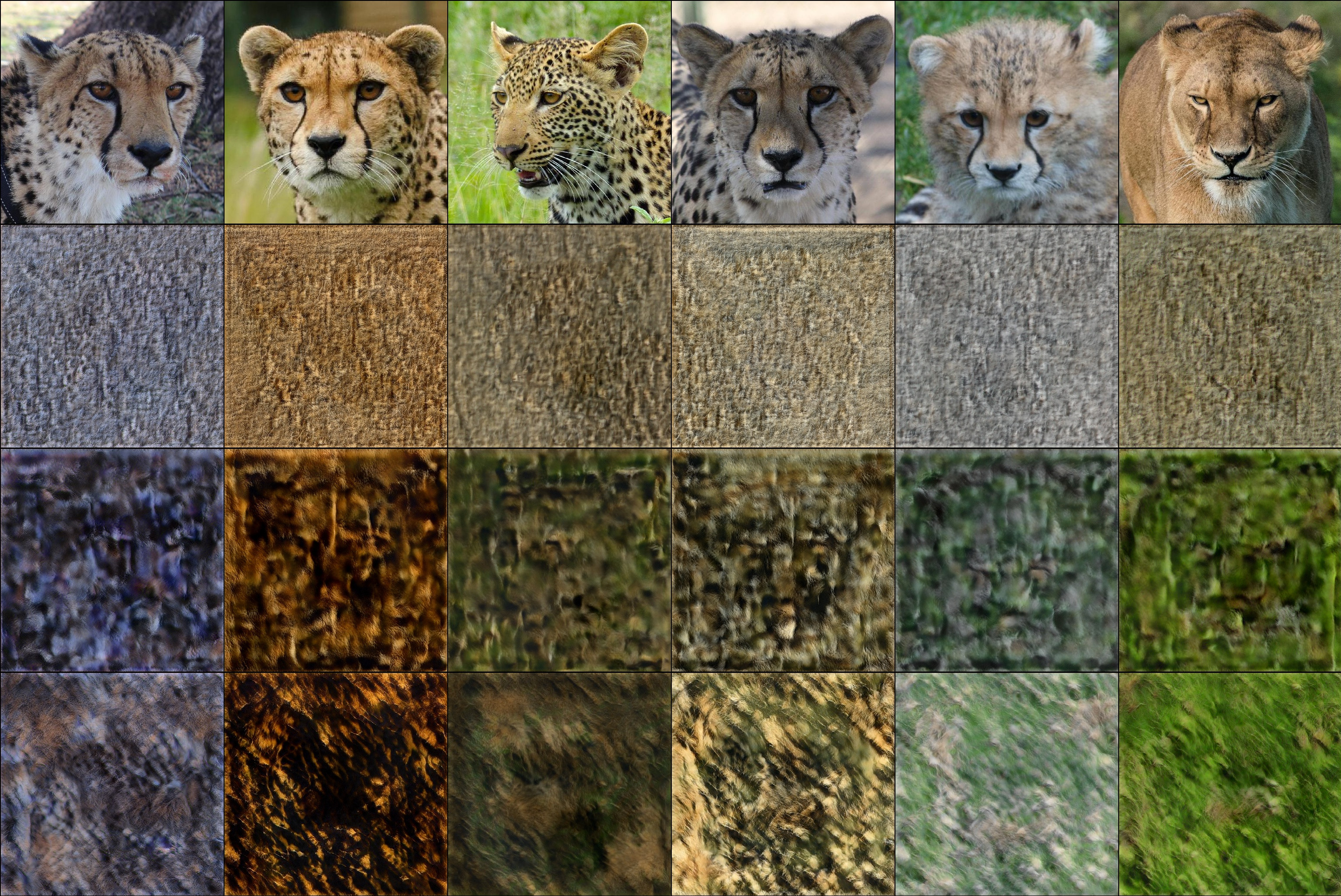}}
        \caption{Reconstruction using StyleGAN2, without using the 1x1 stitching convolution (i.e. directly copying the intermediate output of the feature extractor into the GAN). The rows from top to bottom are: The original dataset image, reconstruction from Layer1, Layer2, Layer3.}
        \label{fig:direct_copy}
    \end{minipage}
}}
\end{center}
\end{figure*}

For BigGAN stitching into deeper layers fails (same number of samplings depth as Layer3 or deeper).
This is visible in \cref{fig:imagenet-samples} in the lower left where the reconstructions from Layer3 result in blue images and do not resemble the original input, and also in the metrics in \cref{tab:afhq-metrics} and \cref{tab:imagenet-metrics} for Layer3 for BigGAN.
BigGAN is class conditional and that conditioning is used as an additional input into the hidden layers.
But for the stitching we ignore it, which likely makes the reconstruction fail for deeper layers (where the network relies on the class conditional input).

\captionsetup[figure]{font=Huge}

Our results further show that StyleGAN2 is more general in earlier layers (same number of samplings depth as Layer1 and Layer2), but becomes more specialized in deeper layers.
That is visible in \cref{tab:imagenet-metrics} where StyleGAN2 is better than BigGAN for reconstructing activations from Layer1 and Layer2 (although this StyleGAN2 was only trained on AFHQ wild and not on the ImageNet data).
But for Layer3 it becomes worse than BigGAN, which was trained on ImageNet, but at a lower resolution of 128x128.  This decline in performance suggests that StyleGAN2 specializes more in the AFHQ wild data here.

While we only use GANs for the reconstruction, it should be possible to use other generative models, e.g., variational autoencoder, in its place.

\subsection{Limitations}

Our method is likely to have problems when the GAN generator cannot understand the concepts of the feature extractor.
%
%In \cref{fig:imagenet-samples}, we use the 1x1 convolutional layers we have trained on the AFHQ wild data to transfer activations from ImageNet data.
%
%The top row shows the original images. The next four rows show reconstructions from our method from Layer1 to Layer4.
%
%We can see that, especially in Layer3, the reconstructions have worse quality than samples for the AFHQ wild data, where the StyleGAN2 generator and 1x1 convolution were trained on.
%
We can see that in \cref{tab:imagenet-metrics} for Layer3 the reconstructions from StyleGAN2 become worse than from BigGAN.
Also class conditioning in the GAN can make problems when stitching into deeper layers of the GAN, since we ignore the class conditioning for reconstruction.

\section{Conclusion and Future Work}
\label{sec:conclusion}

We proposed a novel method to reconstruct activations of a feature extractor by stitching it with a GAN generator.
We extensively evaluated our method and provided evidence that its accuracy is comparable to gradient descent methods while running about two orders of magnitude faster.

We also investigated how the accuracy of the reconstructions is affected when stitching into different layers of the GAN generator.
Our results show that it is a good start to stitch into a layer that has the same number of samplings distance from the output of the GAN generator, as the layer we start from in the feature extractor has from the input.
%
%For 4 out of 6 layers, this choice of target layer to stitch into provided the best results. Only for Layer2 and Layer3 of the ResNet50 network, it was better to stitch into one layer closer to the output of the StyleGAN2 generator.
%
We observe that for most of the layers, the features learned by the feature extractor are compatible with the features learned by a GAN generator, even though the feature extractor and the GAN were trained completely unaware of each other.
Only the deeper layers of a class conditional BigGAN are not compatible with a ResNet50.
Further, we observe that the earlier layers of StyleGAN2 are more general, and the deeper layers become more specialized.

%For future work, it would be interesting to stitch the ResNet50 classifier trained on ImageNet with a GAN that has seen roughly the same variety of concepts.
%
A weakness of the method is that the GAN generator needs to understand the concepts of the feature extractor. Otherwise, the reconstructed image is further away from the original input.
This weakness might, however, be utilized for out-of-distribution (OOD) detection or uncertainty estimation.

\section{Acknowledgements}
%R. Herdt acknowledges the support by the Deutsche Forschungsgemeinschaft (DFG, German Research Foundation) - Projectnumber 459360854. J. Le'Clerc Arrastia, D. Otero Baguer, and M. Schmidt acknowledge the financial support by the German Federal Ministry for Economic Affairs and Climate Action (BMWK) and the European Social Fund (ESF) within the EXIST transfer of research project ``aisencia''.
R.H. is funded by the Deutsche Forschungsgemeinschaft (DFG, German Research Foundation) - project number 459360854. J.L.A., D.O.B., and M.S. acknowledge the financial support by the German Federal Ministry for Economic Affairs and Climate Action (BMWK) and the European Social Fund (ESF) within the EXIST transfer of research project ``aisencia''.

\bibliography{acml23}

\appendix

%\section{First Appendix}\label{apd:first}

\section{Digipath Generator Architecture}\label{apd:GAN Architecture}

\begin{figure*}[ht]
\vskip 0.2in
\begin{center}
\centerline{\resizebox{\linewidth}{!}{
    \centering
    \begin{minipage}{1.4\textwidth}
        \centering
        \includegraphics{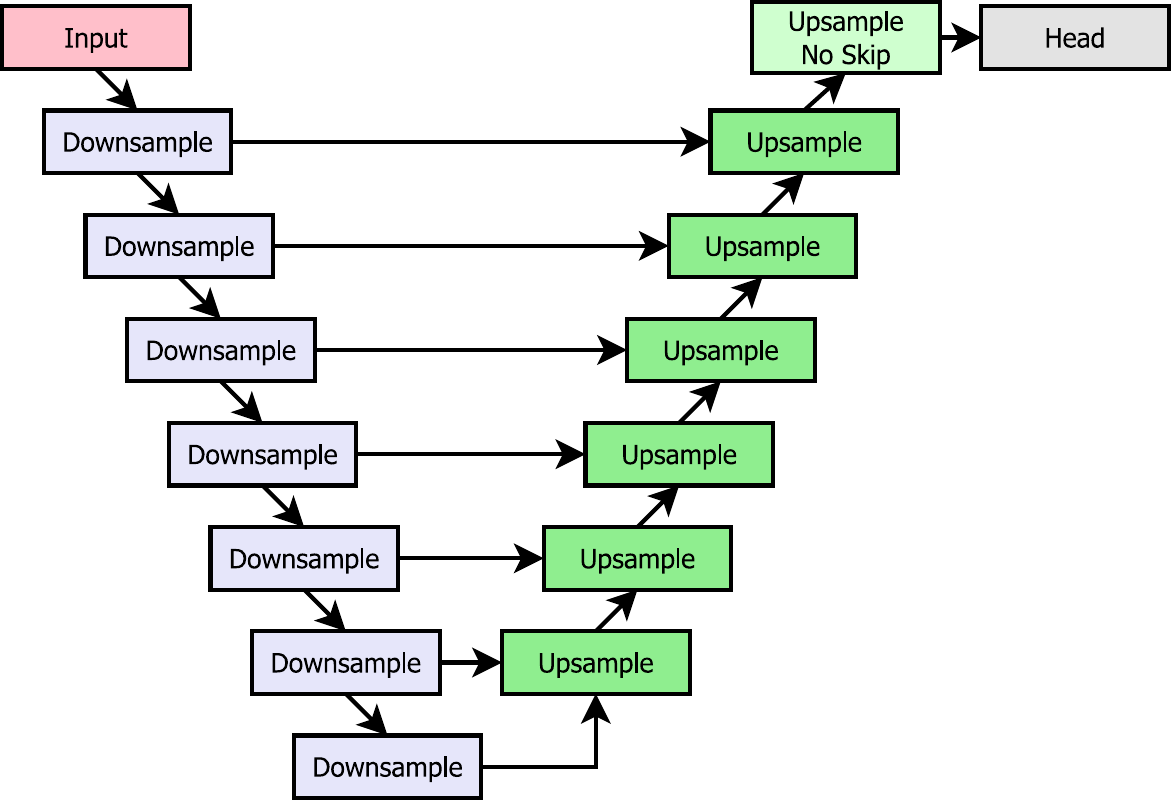}
        \caption{Architecture of the GAN generator we used for the Digipath data. The generated images have a size of 600x600x3.}
        \label{fig:generator-architecture}
    \end{minipage}                  \, \, \, \, \, \, 
    \begin{minipage}{1.4\textwidth}
        \centering
        \includegraphics{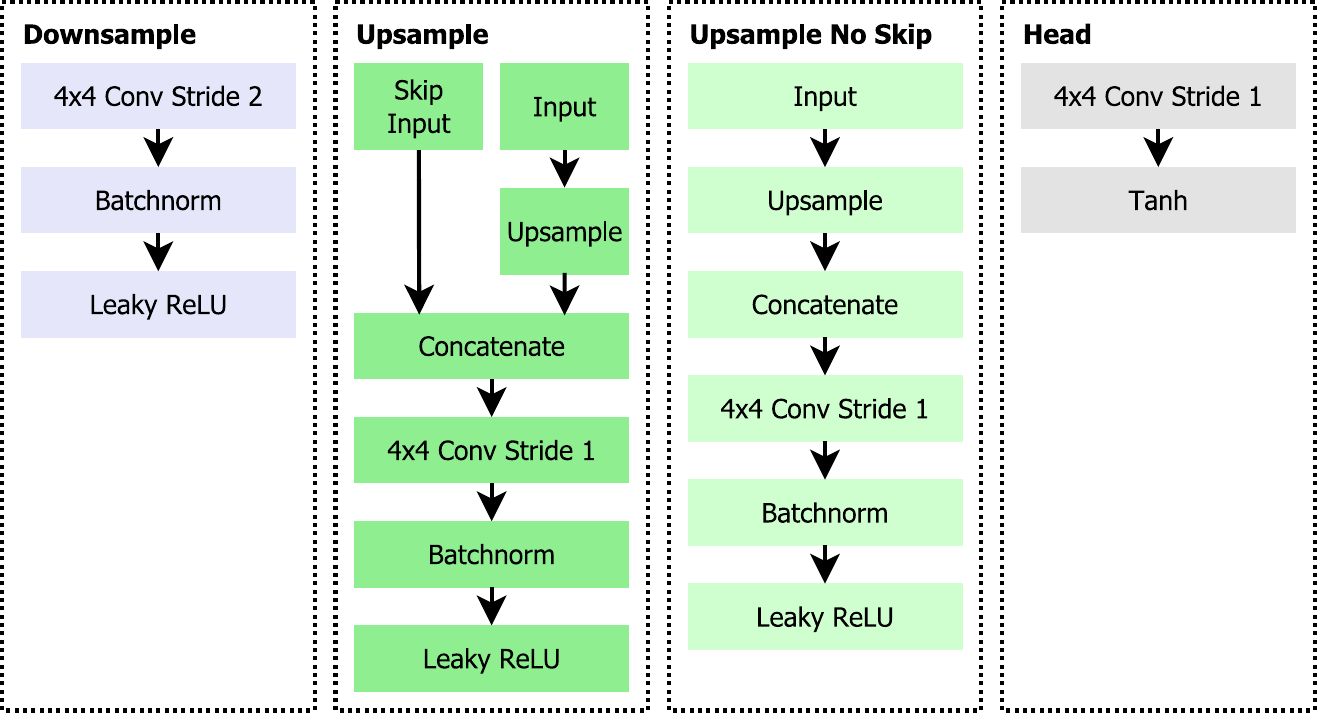}
        \caption{Layers for the GAN generator we used for the Digipath data.}
        \label{fig:generator-architecture-layers}
    \end{minipage}
}}
\end{center}
\end{figure*}

In this section, we describe the architecture of the GAN generator we used for the experiments on the Digipath data.
We use a UNet-like architecture for the generator as shown in \cref{fig:generator-architecture}.
The input is random Gaussian noise of a spatial size of 600x600.
The content for the ''Downsample'', ''Upsample'', and ''Head'' layers is shown in \cref{fig:generator-architecture-layers}.
Each upsample layer has two inputs.
First the input from the skip connection, which is referred to as ''Skip Input'' in \cref{fig:generator-architecture-layers}.
The second input comes from the upsample layer below it, or from the last downsample layer in the case of the first upsample layer.
This second input is referred to as ''Input'' in \cref{fig:generator-architecture-layers}.
It is upsampled via nearest neighbor scaling to the spatial size of ''Skip Input'', and then in the channel dimension concatenated with it.

\captionsetup[figure]{font=Huge}

\end{document}